\documentclass{article}

\usepackage{PRIMEarxiv}

\usepackage[utf8]{inputenc} 
\usepackage[T1]{fontenc}    
\usepackage{hyperref}       
\usepackage{url}            
\usepackage{booktabs}       
\usepackage{amsfonts}       
\usepackage{nicefrac}       
\usepackage{microtype}      
\usepackage{lipsum}
\usepackage{fancyhdr}       
\usepackage{graphicx}       
\graphicspath{{media/}}     

\usepackage{amsmath}
\usepackage{rotating}
\usepackage{xspace}
\usepackage{multirow}

\title{Applying Text Mining to Analyze Human Question Asking in Creativity Research}

\author{
Anna Wróblewska, Marceli Korbin\\
Faculty of Mathematics and Information Sciences,
Warsaw University of Technology\\
 Warsaw, Poland\\
\texttt{anna.wroblewska1@pw.edu.pl}\\
\And
Yoed N. Kenett\\
Faculty of Data and Decision Sciences, Technion - Israel Institute of Technology\\ Haifa, Israel \\
\And
Daniel Dan\\
School of Applied Data Science, Modul University Vienna, 
Austria\\
\And Maria Ganzha\\
Faculty of Mathematics and Information Sciences,
Warsaw University of Technology\\
 Warsaw, Poland\\
\And Marcin Paprzycki\\
Systems Research Institute Polish Academy of Sciences \\ Warsaw, Poland\\
}

\begin{document}
\maketitle

\begin{abstract}
Creativity relates to the ability to generate novel and effective ideas in the areas of interest. How are such creative ideas generated? One possible mechanism that supports creative ideation and is gaining increased empirical attention is by asking questions. Question asking is a likely cognitive mechanism that allows defining problems, facilitating creative problem solving~\cite{kenett2024assessing}. However, much is unknown about the exact role of questions in creativity. This work presents an attempt to apply text mining methods to measure the cognitive potential of questions, taking into account, among others, (a) question type, (b) question complexity, and (c) the content of the answer. This contribution summarizes the history of question mining as a part of creativity research, along with the natural language processing methods deemed useful or helpful in the study. In addition, a novel approach is proposed, implemented, and applied to five datasets. The experimental results obtained are comprehensively analyzed, suggesting that natural language processing has a role to play in creative research.
\end{abstract}


\section{Introduction} \label{sec:intro}
Questions are fundamental tools for acquiring information, resolving confusion, and guiding decision making~\cite{cambridge}. They come in various forms, such as open, closed, hypothetical, rhetorical, leading, and loaded~\cite{differentTypes}. Questions can be ``used'' not only to gain permission or to evaluate competence, but also to explore others’ views, make decisions, or analyze scenarios (this list is, of course, not exhaustive; see also~\cite{15ways,eperel} for a more comprehensive discussion of the topic). Finally, it has been stipulated that questions play an important role in creativity~\cite{akinator,inquiry,aqt,raz2024bridging}. 

Asking questions plays a critical role in the creative process and is an important outcome of the problem construction stage in creative problem solving~\cite{aqt,acar2023,mum1997,reiter1998}. For example, Reiter-Palmon et al. in~\cite{mum1997,reiter1998} have found that people who excel at problem solving tend to restate problems as questions. However, much is still unknown about the specific types of questions asked and their relationship to creative thinking and problem-solving skills. 

Questions are often used to stimulate the recall of prior knowledge, to promote comprehension, and to build critical thinking skills~\cite{tofade2013}. Questions have also been shown to be directly related to creativity. For instance, recent work by Raz et al.~\cite{aqt} found that question complexity was strongly positively related to the question creativity. The authors measured the complexity of the question using Bloom's taxonomy, a widely accepted guideline to design questions of differing levels of cognitive complexity~\cite{adams2015conducting,daud2012}.

However, much is still unknown about the role of asking questions in creativity. This is due to the challenges of empirically assessing various aspects of the questions that usually result in open responses. 
Thus, the research undertaken, the results of which are reported below, faces several limitations at the very beginning. Firstly, there is no established definition of text creativity; while research is growing and suggests creativity-related metrics, no standard (and all-agreed) methods exist. 

Still, at the same time, the theory and practice of automatic natural language processing (NLP) have reached the stage where one could postulate that NLP can be used in research concerning the relations between questions and creativity. In this context, this contribution aims to evaluate the potential of using NLP-based approaches in creativity research. More precisely, the following three research questions will be addressed:
\begin{itemize}
\item \textbf{RQ1}: Can the creativity of questions be measured with NLP techniques?
\item \textbf{RQ2}: How can one measure whether, and how, various types, or characteristics of a question influence creativity?
\item \textbf{RQ3}: How does the question lead to a more creative answer?
\end{itemize}

In general, throughout this contribution, current (end 2024) NLP techniques are explored to help understand how to assess, or classify, questions in the context of their relations to creativity\footnote{Our source code is available under the link \url{https://github.com/Z-Xbeova/Creativity-Research-Question-Mining-MSc}.} Here, it should be noted that this task is not only difficult but also quite ambiguous (even for humans). Therefore, while it may be difficult to fully answer the research questions that have been posed, the more realistic goal is to determine which techniques, why, and how, are worthy further considerations in future research.

%
%

In this context, in Section~\ref{sec:rel_research}, previous research is introduced as a background to the reported work. Section~\ref{sec:data_sources} presents data sources used in the experiments. Section~\ref{sec:experiments} introduces proposed experiments, to be used to answer the three research questions. The results are then described, in Section~\ref{sec:keyresults}, and further discussed in Section~\ref{sec:discussion}, eventually giving way to conclusions in Section~\ref{sec:conclusions}.

\section{Related Research} \label{sec:rel_research}
Let us start by summarizing the past research connected to the conducted research, including basic tools that can be used to classify questions, the role of question-asking and question analysis in the creative process, and examples of the complexity measures and of useful natural language processing methods.

\subsection{Characteristics of Questions} \label{sec:char}
The first issue that should be considered is the state of the art of understanding the characteristics of the questions. Questions can be categorized in various ways, such as open-ended (e.g., “What is a computer?”), closed (e.g., “Is this a computer?”), hypothetical (e.g., “What would you do if your computer stopped working?”), rhetorical (e.g., “Who wouldn’t want a computer?”), leading (e.g., “Computers are beneficial, aren’t they?”), and loaded (e.g., “Why don’t you own this computer yet?”)~\cite{differentTypes}. These categories of questions vary by the range of answers, the knowledge sought, and complexity. The last one can be evaluated using Bloom’s Taxonomy~\cite{taxonomy}, introduced by Benjamin Bloom in 1956 and revised in 2001. This taxonomy was designed to help educators to structure the learning objectives and the assessments, by organizing them into six hierarchical categories in the cognitive domain~\cite{taxonomy,stems,taxonomyWebpage}.
\begin{itemize}
\item \textbf{Knowledge} – recalling information, e.g., “What is a computer?”
\item \textbf{Comprehension} – understanding content, e.g., “How does a computer work?”
\item \textbf{Application} – using knowledge in practice, e.g., “How can a computer facilitate one’s life?”
\item \textbf{Analysis} – examining relationships, e.g., “What are the effects of using a computer depending on workflow?”
\item \textbf{Synthesis} – creating new ideas, e.g., “How can computers be made to work more effectively?”
\item \textbf{Evaluation} – assessing information, e.g., “How beneficial are computers for the world?”
\end{itemize}

The revised taxonomy updated the names of these categories, as active verbs: remember, understand, apply, analyze, evaluate, and create~\cite{taxonomy}. The new version emphasizes the higher-order thinking skills, placing ``create'' above ``evaluate''. It is worth noting that the affective and the psychomotor domains have been also considered~\cite{krathwohl,simpson}. However, this work focuses only on the cognitive domain.

\subsection{Question Asking in the Creative Process}\label{sec:QA}
Creativity can be understood as using existing ideas to search for new ones. It involves many cognitive processes, and is associated with adaptability, flexibility, survival, and evolution, in complex environments~\cite{sonophilia}. The cognitive power of asking questions has been suggested to significantly enhance the creative process. According to~\cite{sonophilia}, complex questions are more likely to generate detailed ideas or concepts, as their formulation shapes how a problem is identified. Further, question-asking improves problem-finding, which, in turn, boosts creativity.

A study titled ``A Mirror to Human Question Asking: Analyzing the Akinator Online Question Game''~\cite{akinator} explored the popular online game Akinator, where a genie guesses a character through a series of questions. The researchers analyzed the game’s questioning strategies, using topic modelling and clustering, noting parallels between the Akinator’s method and the human question-asking. This way, the study was the first step towards determining how questions can improve the creative process.

Although the role of questions in creativity remains underexplored, research in this area is growing. Kenett et al.~\cite{inquiry} describe question complexity as linked to the creativity of the person that asked them. They found that complex questions often lead to deeper analysis. However, evaluating complexity requires considering factors like context and cognitive processes.

Bloom’s Taxonomy is frequently used to classify questions by abstraction and answer variety. In a study entitled ``The Role of Asking More Complex Questions in Creative Thinking''~\cite{aqt}, participants were tasked with creating unusual questions about common objects, which were then evaluated using Bloom’s Taxonomy. The results indicated that more complex questions, as defined by the taxonomy, could have been correlated with higher creativity, supporting its use in assessing the question complexity.

\subsection{Question Analysis in Automated Tasks}\label{sec:QAnalysis}
Question analysis is helpful in the process of answer generation, which is crucial to automatize the increasing number of questions. This challenge was addressed in ``What makes us curious? Analysis of a corpus of open-domain questions''~\cite{wtc}, where the authors analyzed a dataset from \textit{Project What If} (run by \textit{We the Curious}~\cite{whatif}). They aimed to develop an AI tool not only for answering questions but also for determining topics and detecting question equivalence. The proposed tool, called QBERT, was built by fine-tuning an S-BERT model~\cite{sbert}. One of the achievements of the study (\cite{wtc}) was to automatically separate the factual (e.g., “Has a computer facilitated your life?”) from the counterfactual questions (e.g., “If computers didn’t exist, how would the world function today?”), with the latter requiring more detailed responses. This finding served mainly as a hint towards what characterizes a creative question.

Another large-scale question analysis is discussed in ``Text Mining of Open-Ended Questions in Self-Assessment of University Teachers''~\cite{ecuador}. Here, the Latent Dirichlet Algorithm (LDA) topic modeling~\cite{lda_ml} was applied to process the survey data. The authors visualized topic clusters, identifying strategies such as practical learning, use of technology, and teamwork, to improve student retention. They considered visualization as a highly important aspect of text mining. The methodology helped ascertain teachers' main strategies to improve student retention.

Exam question classification, using Bloom's taxonomy, has also been explored. Here, Gani et al.~\cite{examq} proposed a model combining a CNN and RoBERTa~\cite{roberta}, which outperformed earlier models~\cite{eq1,eq2,eq3,eq4,eq5,eq6} accuracy-wise in the Bloom's Taxonomy level classification task, including RNN-based approaches. The authors also proposed a dataset of exam questions and task classified according to the Bloom's taxonomy, which we utilized in this paper.

\subsection{NLP Techniques for Question Assessment}\label{sec:nlp}
%
There are several ways to assess text complexity (including questions), with the most common approaches focusing on lexical diversity, or readability, to more advanced topic modelling and embedding-based approaches. Here, let us introduce the most popular techniques, which are also further explored in this work.

\subsubsection{Lexical Diversity}
Lexical diversity refers to the range of different words used in a text. It is widely applied in fields such as neuropathology and data mining to assess communication skills, in both written and spoken contexts. While some popular indices, for measuring lexical diversity, have been criticized for their sensitivity to the text length or homogeneity, the latest methods are addressing these issues~\cite{ld}.

Currently, several key methods are used to measure lexical diversity. A straightforward approach is to analyze the length of the longest word in the text~\cite{voceval}. However, it is not very popular. A more common measure is the type-token ratio (TTR), which calculates the proportion of unique words to the total number of words~\cite{ttr}. The corrected type-token ratio (CTTR) adjusts this measure by accounting for the length of the text. It is calculated as:
\[
\text{CTTR} = \frac{\text{number of unique words}}{\sqrt{2 \times \text{number of words}}}.
\]
Another diversity measure is Simpson’s diversity index, also called the D metric, which evaluates the frequency of word repetition:
\[
D = \sum_{i=1}^{V(N)} f_v(i, N) \frac{i(i-1)}{N(N-1)},
\]
where $V(N)$ is the number of unique words, $N$ is the total number of words, and $f_v(i, N)$ is the frequency of words occurring $i$ times~\cite{lexrich}. Content uniqueness can also be measured by analyzing the number of repeated $n$-grams, providing insight into the diversity and originality (see,~\cite{marketing} for more details). 

In what follows, metrics developed specifically for creativity research, have been applied. These include: divergent semantic integration (DSI)~\cite{dsi}, a measure which calculates the metric distance between every pair of words in the text. Here, cosine distance between word embeddings isa applied. Furthermore, maximum associative distance (MAD)~\cite{mad} -- a measure based on the semantic distance of the word that is the furthest from the prompt (question) word, is used.

\subsubsection{Text Readability}
Text readability is the combination of elements that affect a reader’s ability to comprehend, process, and engage with a text. It depends on factors such as the style, format, context, and the reader’s background~\cite{readable}.
Readability is being assessed using various tools. For instance, basic statistics, such as mean sentence length and average syllables per word, are common starting points. A widely used measure is the Flesch reading-ease score (FRES), which calculates readability based on the sentence length and the syllable count:
\[
FRES = 206.835 - 1.015 * \text{average sentence length} - 84.6 * \frac{\text{number of syllables}}{\text{number of words}}
\]
Here, higher FRES values indicate easier, more conversational text, while lower scores represent a more difficult material, typically requiring a higher education level to understand~\cite{flesch}. Another metric, the Flesch-Kincaid grade level (FKGL), is a variation of FRES that correlates text complexity with the U.S. school grade levels:
\[
FKGL = 0.39 * \text{average sentence length} + 11.8 * \frac{\text{number of syllables}}{\text{number of words}} - 15.59
\]
The Automated Readability Index (ARI) is another text readability measure that factors-in both sentence length and word length, to estimate the readability:
\[
 ARI = 0.5 * \text{average sentence length} + 47.1 * \text{average word length} - 21.34
\]
These metrics offer a detailed assessment of text readability, helping to evaluate how accessible a text is to the different audiences.

\subsubsection{Sentence Embedding}
Let us now direct attention to the NLP approaches pertinent to the research reported in what follows. Sentence embedding is a technique, in NLP that represents text as a fixed-length numeric vector, while preserving its semantic meaning. It has been developed by averaging word embeddings, using pre-trained models, and applying neural networks.

Averaging word embeddings involves embedding each word in the text (using algorithms like word2vec~\cite{w2v}, GloVe~\cite{glove}, or ELMo~\cite{elmo}) and then averaging these embeddings. This approach is practical for resource-limited tasks, with simple sentence structures.  

\subsubsection{Transformers}
Pre-trained transformer models are better suited for capturing the full context of a text. Transformers, introduced by Vaswani et al. in 2017~\cite{transformer}, revolutionized Natural Language Processing (NLP) by eliminating the need for recurrence, or convolution and, instead, relying on an encoder-decoder structure with attention mechanisms. These mechanisms map queries and key-value pairs to outputs, allowing for the modelling of global dependencies between inputs and outputs. 

Here, a key development is BERT (Bidirectional Encoder Representations from Transformers)~\cite{bert}, which uses bidirectional context to understand the meaning of sentences, by considering words both before and after the target word. BERT is pre-trained using masked language modelling and next-sentence prediction, and fine-tuned for specific downstream tasks, marking it as a cornerstone in modern NLP. It also introduced transfer learning~\cite{transfer} to NLP, allowing models to be reused across different tasks. Notable modifications of BERT include Sentence-BERT (S-BERT) for sentence similarity~\cite{sbert}, RoBERTa, which refines pre-training~\cite{roberta}, and DistilBERT, a lightweight version with fewer parameters but similar performance~\cite{distilbert}. MiniLM~\cite{minilm} compresses transformer models, reducing parameters while retaining performance by using self-attention mechanisms and scaled dot-product values. MPNet~\cite{mpnet} combines the strengths of BERT and XLNet~\cite{xlnet}, utilizing both bidirectional context and permuted language modelling to capture token dependencies and sentence structure more effectively. Additionally, the Generative Pre-trained Transformer (GPT) series, introduced by Radford and Narasimhan~\cite{gpt1}, applies the transformer architecture to generate human-like text responses. The GPT models, including GPT-2~\cite{gpt2}, GPT-3~\cite{gpt3}, GPT-4~\cite{gpt4} (and its variants, released almost weekly, in the second half of 2024), have progressively enhanced performance, setting the stage for large-scale language models and gaining popularity with applications like ChatGPT~\cite{forbes_gpt,forbes_io}, which is widely recognized for its sophisticated and detailed responses. These transformer-based embedding representations, e.g. for questions, can potentially better explain and also show the creative characteristics of the questions.

\subsubsection{Topic Modelling}
Topic modelling is another technique, which can be used to explore different kinds of questions related to different topics. Here, related techniques have the ability to identify themes within large datasets~\cite{blei_lafferty}. Specifically, {Latent Dirichlet Allocation (LDA)}~\cite{lda_ml} is a probabilistic model that represents documents as mixtures of topics. It was first introduced in the population genetics area in 2000~\cite{popgen_lda} and used in Machine Learning since 2003~\cite{lda_ml}. 
The model has a hierarchical, three-level structure, modelling every document as a combination of provided topics, which are conversely viewed as mixtures of words. It attempts to recognize a latent structure, underlying a collection of documents, basing it on topics and their word distributions. Every document is assigned a distribution of topic probabilities as its representation.
Another, more sophisticated, algorithm for topic discovery is {Seed-Guided Algorithms}~\cite{seedtm}, also known as \textit{SeedTopicMine}. It is a topic modelling method, based on seed words, which are the most representative of certain topics, usually chosen beforehand by experts to help determine the topic of a document. The algorithm is described as a combination of supervised and unsupervised machine learning, with the supervised branch being especially represented by seed words. This approach is meant to fix a problem specific to LDA, where retrieved topics are semantically general and do not necessarily align with users' specific interests. 

In the reported research, these NLP-based techniques have been applied to the texts of questions, or to the the combinations of questions and answers, utilizing the datasets described in the following section.

\section{Data Sources}\label{sec:data_sources}

Datasets with different kinds of questions were collected for this work. They were not designed specifically to measure creativity. Instead, they were focused more on fact-checking rather than on the relations between questions and answers and on the creative thinking. These datasets, comprising question texts and, most often, also question and answer texts, which we further explored, are as follows -- open datasets: Exam Question Dataset, R. Tatman’s Question-Answer Dataset, Question-Answer Jokes, the Stanford Question Answering Dataset, and proprietary transcripts from the Slovenian Press Agency (STA). Table~\ref{dataset_table} provides a summary comparison of these datasets.
\begin{table}[htp]
\centering
\resizebox{\textwidth}{!}{%
\begin{tabular}{llllll}
\hline
\textbf{Dataset} & \textbf{\begin{tabular}[c]{@{}l@{}}Number of\\ Questions\end{tabular}} & \textbf{\begin{tabular}[c]{@{}l@{}}Has\\ Answers?\end{tabular}} & \textbf{\begin{tabular}[c]{@{}l@{}}Mean Number\\ of Characters\\ in a Question\end{tabular}} & \textbf{\begin{tabular}[c]{@{}l@{}}Mean Number\\ of Words\\ in a Question\end{tabular}} & \textbf{\begin{tabular}[c]{@{}l@{}}Context of Usage\end{tabular}} \\ \hline
\begin{tabular}[c]{@{}l@{}}Exam Question\\ Dataset\end{tabular} & 2,522 & No & 96 & 16 & \begin{tabular}[c]{@{}l@{}}Bloom's Taxonomy\\ levels\end{tabular} \\ 
\begin{tabular}[c]{@{}l@{}}R. Tatman’s\\ Q-A Dataset\end{tabular} & 2,150 & Yes & 54 & 10 & \begin{tabular}[c]{@{}l@{}}Difficulty levels,\\ answers to questions\end{tabular} \\ 
\begin{tabular}[c]{@{}l@{}}Question-Answer\\ Jokes\end{tabular} & 38,266 & Yes & 49 & 10 & \begin{tabular}[c]{@{}l@{}}Answers to questions,\\ unconventional content\end{tabular} \\ 
Stanford Dataset & 104,564 & Yes & 60 & 11 & Answers to questions \\ 
STA Interviews & 3,780 & Yes & 648 & 115 & \begin{tabular}[c]{@{}l@{}}interviews - timestamps, \\speaker information, \\questions \& answers\end{tabular} \\ \hline
\end{tabular}%
}
\caption{Datasets utilized at this work}
\label{dataset_table}
\end{table}

The Exam Question Dataset consists of questions, with labels from Bloom's taxonomy, and it best suits the experiments, because the taxonomy levels are the most related to the creativity notion (as described above). R. Tatman’s Dataset, Question-Answer Jokes, and the Stanford Dataset are more tangential. However, they consist of questions and answers, and they were selected for their quality, and ease of access. Moreover, of importance was that they cover diverse topics. The Slovenian Press Agency transcripts were used in the last experiment, in which the goal was to assess the flow of a complete interview, and how the subsequent questions and answers (and their flow) relate to the creativity.

Noteworthy, we assumed that in the datasets with the question-answer pairs, such as R. Tatman’s Question-Answer Dataset, Question-Answer Jokes, and Stanford Question Answer Dataset, all responses adhere to the question’s topic. Answers that deviate too far may make the question seem creative but are unlikely to be helpful in the creative process itself. Nevertheless, taking into account the fact that some brianstorming approaches may involve ``off the wall'' questions and answers, this aspect merits further, independent, exploration.

The datasets were preprocessed, with the questions and answers transformed into lowercase and tokenized. The Exam Question Dataset~\cite{examq} contains 2,522 exam questions labelled with Bloom’s Taxonomy levels, averaging 96 characters and 16 words per sentence. The R. Tatman’s Question-Answer Dataset~\cite{rtatman} includes 2,150 unique questions generated from Wikipedia articles, with difficulty levels assigned by both the questioners and the answerers. On average, each question is 54 characters long, with 10 words, and most answers are short. Question-Answer Jokes~\cite{jiriroz} contains 38,266 jokes in a question-answer format, retrieved from Reddit’s r/Jokes. An average question in this dataset has 49 characters and 10 words, with short answers. The Stanford Question Answering Dataset (SQuAD)~\cite{stanfordu} includes 104,564 questions related to over 500 Wikipedia articles, averaging 60 characters and 11 words per question, paired with longer answers. Transcripts from the Slovenian Press Agency (STA) contain 244 interview transcripts, with questions averaging 648 characters and 115 words. This dataset includes a sequence of questions and answers, as well as timestamps and partial speaker labelling, although 84\% of the rows lack speaker information.

\section{Experimental Setup}\label{sec:experiments}
%
The aim of this work is to check if and how the NLP-based methods can help determine and measure question creativity, thus potentially automating (at least some aspects of) creativity research. To explore this, measures to assess creativity were developed, and their relevance tested, but still with a limited conceptual creativity assessment. While the reported work focuses on question creativity, previous studies~\cite{inquiry,aqt} have concentrated on question complexity. Although not identical, the two traits are suggested to be correlated, allowing one to reinterpret complexity measures as creativity measures.

To experimentally explore the relations between questions (and answers) and creativity, four experiments were designed:
\begin{enumerate}
\item Taxonomy Assessment: question exploration with Bloom's Taxonomy levels vs. readability and lexical diversity metrics.
\item Answer-Based Assessment: question and answer exploration with text-based metrics.
\item Topic Study: the two above explorations, but divided into different topics.
\item Interview Study: the above explorations, applied within the flow of the whole interview, i.e. subsequent questions and answers.
\end{enumerate}. 

The first two experiments focused on Bloom's Taxonomy Assessment and the Answer-Based Assessment. These experiments aimed to quantify question creativity, with various question features, such as the number of characters, words, common word percentage, diversity, and readability indices.

The Taxonomy Assessment and the Answer-Based Assessment experiments directly addressed research question RQ1 by measuring questions with NLP techniques. The Taxonomy Assessment further explored RQ3, providing insights into how certain features of questions foster creativity. 
The Answer-Based Assessment addressed RQ2,by jointly analysing questions with their corresponding answers. 

The subsequent two experiments -- the Topic Study and the Interview Study -- explored additional factors, like question topics and their occurrence in the interviews, addressing both RQ2 and RQ3.
Below, we present more details on how each of the four experiments was conducted. 

\subsection{Taxonomy Assessment}

The goal of the Taxonomy Assessment experiment was to evaluate if, and how well, Bloom’s Taxonomy can measure question creativity. Bloom’s Taxonomy is one of the earliest frameworks for classifying questions, and prior research suggests its relevance to the creativity~\cite{inquiry,aqt}. 

We replicated and trained the classification model from~\cite{examq}, which assigns Bloom’s Taxonomy levels to the questions, outperforming previous models~\cite{examq,eq1,eq2,eq3,eq4,eq5,eq6}.
We used the Exam Question Dataset (the same as in~\cite{examq}) for training and testing our model. First, questions from the Exam Question Dataset were transformed into token embeddings using RoBERTa \textsubscript{base}, then passed through a convolutional neural network (CNN) to predict Bloom’s Taxonomy levels. Then, the trained model was applied to all our question-answer open datasets (including R. Tatman’s Question-Answer Dataset, Question-Answer Jokes, and Stanford Question Answering Dataset). Then we calculated the complexity measures: TTR, D metric, Flesch reading-ease score (FRES), and the Automated Readability Index (ARI), etc., and compared them against the Bloom’s Taxonomy levels.

\subsection{Answer-Based Assessment}
The Answer-Based Assessment aimed to examine how the questions prompt creative answers. We developed 
and tested four creativity measures for the question-answer pairs, including all the metrics proposed in Section~\ref{sec:nlp}. Here, these metrics were utilized to assess the combined question-answer text rather than the answer alone, as creative answers should introduce new information that alters the conversation context. 
Our analysis compared these metrics across datasets: R. Tatman’s Question-Answer Dataset, Question-Answer Jokes, and the Stanford Question Answering Dataset.

In the Answer-Based Assessment, the information and the content uniqueness metrics were computed for each question-answer pair. The cosine distance between the question and the concatenated question-answer embeddings was used to compute the information, while content uniqueness was derived from the bigram comparison between the question and the combined text. Additionally, DSI and MAD metrics were applied, and results were compared with those from the Taxonomy Assessment.

\subsubsection{Metrics adjustment to question-answer pair as input data.}
For each question-answer pair, we computed the following metrics: information metric, content uniqueness metric, DSI, and MAD with computation procedures as follows. 
\linebreak
Information metric:
    \begin{itemize}
        \item The question text string is transformed into a vector $V1$ with a sentence embedding; 
        \item The concatenation of the question and answer text strings is embedded as a vector $V2$;
        \item The information metric is computed as the cosine distance between the vectors $V1$ and $V2$.
    \end{itemize}
Content uniqueness metric:
    \begin{itemize}
        \item Two sets of bigrams (where bigram is a sequence of two tokens) are constructed: one ($S1$) from the question text string, the other ($S2$) from the concatenation of the question and answer text strings;
        \item $n_{ab}$ is computed as a number of all possible bigrams in $S2$;
        \item $n_{db}$ is computed as a number of bigrams from $S2$ that appear more than once in $S2$, or appear also in $S1$;
        \item The content uniqueness metric is computed as $1-\frac{n_{db}}{n_{ab}}$.
    \end{itemize}

\noindent 
Additionally, we computed the DSI \cite{dsi} and MAD \cite{mad} metrics; In its original form, MAD requires a single prompt word and the answer text as inputs. Here, however, every word in the question (excluding stopwords) is used as a prompt word to compute MAD, and the maximum value achieved across the question is returned as a result.

For the sentence embedding task, we use three models from the SentenceTransformers framework \cite{sbert,sbertnet}:
\begin{itemize}
    \item \texttt{all-distilroberta-v1}, based on DistilRoBERTa -- a distilled version of the RoBERTa\textsubscript{base} model \cite{roberta} \cite{distilbert} (size: 290 MB);
    \item \texttt{all-MiniLM-L12-v2}, based on the MiniLM model \cite{minilm}  (size: 120 MB);
    \item \texttt{all-mpnet-base-v2}, based on the MPNet model \cite{mpnet} (size: 420 MB).
\end{itemize}

\subsection{Topic Study}
The Topic Study investigated how the question creativity varies by topic, using topic modelling tools -- Latent Dirichlet Allocation (LDA) and SeedTopicMine. Questions weregrouped into topics, either manually for R. Tatman’s Dataset, or through topic modelling for the Question-Answer Jokes and Stanford Question Answering Dataset. 
We analyzed question creativity across topics, comparing results with those from the Taxonomy Assessment and Answer-Based Assessment experiments.

\subsection{Interview Study}

Finally, the Interview Study explored the relationship between the creativity and the timing of the questions in interviews using the STA interviews dataset. Eight fully labelled long interviews (with also longer questions) were selected, and question-answer pairs were analyzed based on Bloom’s Taxonomy levels and answer-based creativity metrics. The experiment visualizes creativity as a timeline (sequence), plotting the DSI metric against time and marking Bloom’s taxonomy levels. 

In the Interview Study, we analyzed eight STA interviews, examining how question creativity fluctuated throughout the interview timeline. Table \ref{sta_e4} presents the interview dataset statistics.
\begin{table}[ht]
\centering
\begin{tabular}{lcccc}
\hline
\textbf{Nr.} & \textbf{\begin{tabular}[c]{@{}l@{}}Statements: Q or A\\ -- Row Count\end{tabular}} & \textbf{\begin{tabular}[c]{@{}l@{}}Interview \\Time \\(min:sec)\end{tabular}} & \textbf{\begin{tabular}[c]{@{} l@{}}Avg. Question \\Chars\end{tabular}} & \textbf{\begin{tabular}[c]{@{}  l@{}}Avg. Answer \\Chars\end{tabular}} \\ \hline
1 & 14 & 35:36  & 516.5  & 1252.4  \\ 
2 & 11 & 24:34  & 969.7  & 638.3   \\
3 & 8  & 46:33  & 829.1  & 4123.9  \\ 
4 & 11 & 39:20  & 646.6  & 2092.5  \\ 
5 & 5  & 18:26  & 478.8  & 1352.6  \\ 
6 & 16 & 47:07  & 733.8  & 1170.8  \\ 
7 & 16 & 51:00  & 566.7  & 2068.7  \\ 
8 & 4  & 9:43   & 653.3  & 1025.3  \\\hline
\end{tabular}
\caption{Statistics of the STA interviews used in the Interview Study}
\label{sta_e4}
\end{table}

\section{Key Results}\label{sec:keyresults}

The selected results of each experiment are described in the following sections.

\subsection{Experiment 1: Taxonomy Assessment}

\subsubsection{Classification model training process}

\begin{figure}[htp]
  \centering
  \includegraphics[width=0.6\textwidth]{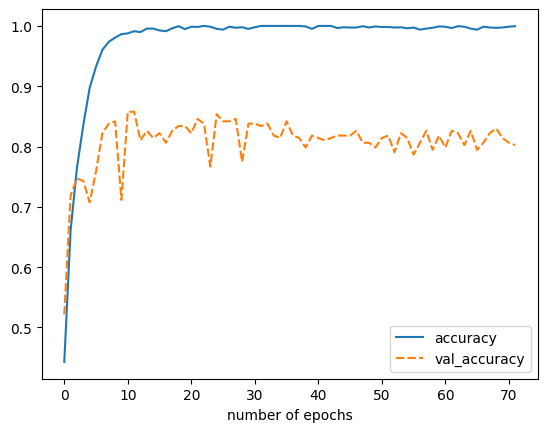}
  \caption{Training and test accuracy during the training classification model to recognize Bloom's Taxonomy levels}
  \label{e1_model_accuracy}
\end{figure}

As noted, our classification model to recognize Bloom's Taxonomy Level was trained on the Exam Question Dataset. Figure~\ref{e1_model_accuracy} shows the accuracy over the course of the training. The model achieved near-perfect accuracy on the training set and about 80\% accuracy on the test set by the 10th epoch. This is similar to the performance reported in the source paper~\cite{examq}. Subsequently, we applied this model to other datasets to differentiate the questions into Bloom's Taxonomy levels. 

\subsubsection{Model prediction explanation results}

\begin{figure}[!htb]
    \centering
\includegraphics[width=0.8\textwidth]{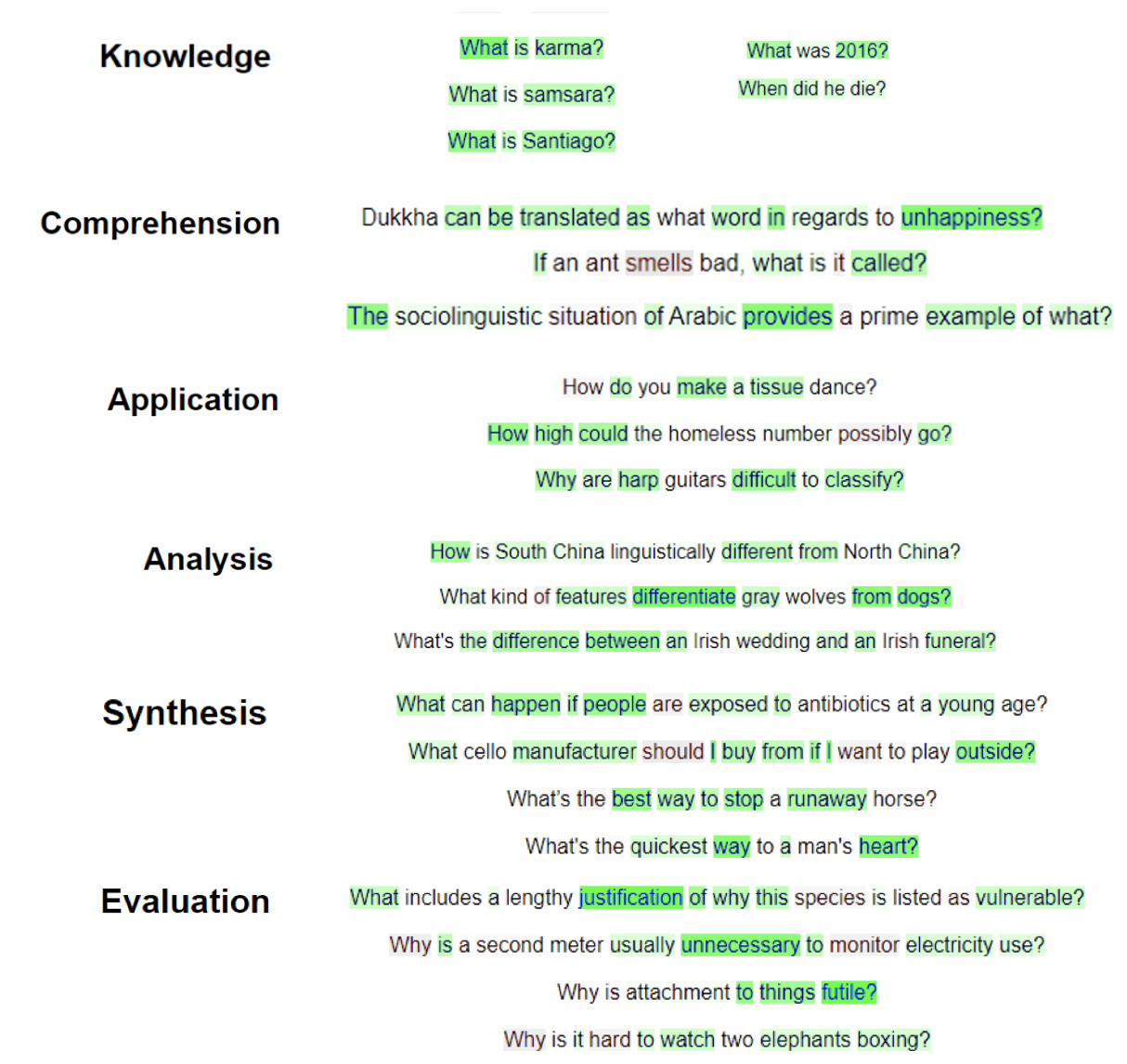}
    \caption{Results of model prediction explanations on the  questions from each Bloom's Taxonomy level}
    \label{e1_examples_lime1}
\end{figure}

Figure~\ref{e1_examples_lime1} provides explanations for all the presented questions prepared with the LIME~\cite{lime} package. Words marked as the most significant in the classification are marked with a green background, while the least likely words for the assigned level are on a grey background. Examples of the most indicative words for every Bloom's Taxonomy level are:
\begin{itemize}
    \item Knowledge: \textit{what}, \textit{who}, \textit{in}, proper names, e.g. Santiago;
    \item Comprehension: \textit{provide}, \textit{translated}, \textit{meaning};
    \item Application: \textit{about}, \textit{draw}, \textit{make}, \textit{difficult};
    \item Analysis: \textit{compare}, \textit{difference}, \textit{how};
    \item Synthesis: \textit{can}, \textit{way}, \textit{happen};
    \item Evaluation: \textit{why}, \textit{justification}, \textit{unnecessary}.
\end{itemize}

\subsubsection{Distribution of Bloom’s Taxonomy levels}
\begin{figure}[htp]
  \centering\includegraphics[width=\textwidth]{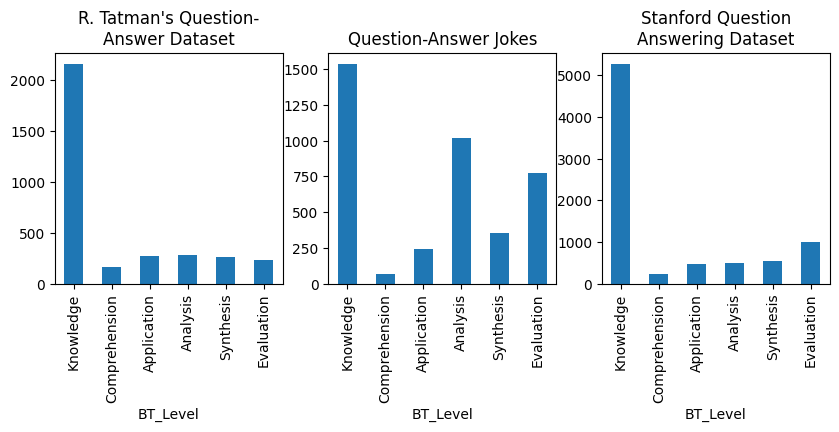}
  \caption{Bloom’s Taxonomy levels distribution for three datasets}
  \label{e1_bt_levels}
\end{figure}

The Bloom’s Taxonomy classification model was applied to three datasets: R. Tatman’s Question-Answer Dataset, Question-Answer Jokes, and Stanford Question Answering Dataset. Figure~\ref{e1_bt_levels} shows the distribution of predicted taxonomy levels. In all datasets, the Knowledge level dominates, particularly in R. Tatman’s and the Stanford's datasets. Analysis and Evaluation are more prevalent in the Question-Answer Jokes dataset. Comprehension is the least frequent level across all datasets.

The prevalence of the Knowledge level in the R. Tatman and Stanford datasets may result from their factual nature, which typically focuses on testing the knowledge. In contrast, Question-Answer Jokes contain more questions classified as referring to Analysis and Evaluation levels, likely due to the nature of humour, which indicates more complex and possibly creative texts.

\subsubsection{Metrics within datasets}
Table~\ref{tab:e1_measures_vs_blooms} presents the mean and the standard deviation and ranges of our metrics for each dataset and a more detailed analysis related to Bloom's Taxonomy levels for the two datasets. The Knowledge questions tend to be the simplest across all datasets (see CTTR and FKGL metrics), while the Comprehension and Analysis questions tend to show higher complexity in some cases. However, no clear correlation between the taxonomy level and the other metrics was observed, while the differences between levels were subtle.
%
\begin{sidewaystable}
\centering
\addtolength{\tabcolsep}{+3pt}    
\begin{tabular}{rccccccccc}
\hline
Dataset  & \# Chars & \# Words  & \begin{tabular}[c]{@{}l@{}}Common \\Words \end{tabular}& TTR   & CTTR & D-Metric  & FRES  & FKGL   & ARI  \\

\multicolumn{10}{c}{Mean values}\\
RTD & 52 ± 25  & 9.9 ± 4.5 & 0.49 ± 0.14  & 0.98± 0.052  & 2.1± 0.36  & 0.0044 ± 0.01 & 27 ± 40 & 22 ± 11 & 250 ± 39 \\ 
 QAJD &  48 ± 15 &  10 ± 2.9 &  0.6 ± 0.14 &  0.97 ± 0.054 &  2.2 ± 0.29 &  0.0057 ± 0.012 &  52 ± 28 &  18 ± 7 &  220 ± 31 \\
SQAD &  60 ± 20 &  11 ± 3.5 &  0.54 ± 0.12 &  0.98 ± 0.045 &  2.3 ± 0.32 &  0.0039 ± 0.0081 &  22 ± 35 &  25 ± 9.2 &  260 ± 34 \\
\hline
\multicolumn{10}{c}{Range values}\\
RTD & 15 ÷ 270    & 4 ÷ 50   & 0 ÷ 0.89  & 0.65 ÷ 1 & 1.4 ÷ 4.1 & 0 ÷ 0.073 & -220 ÷ 140 & -1.9 ÷ 110 & 130 ÷ 410\\
QAJD &  9 ÷ 100 &  2 ÷ 23 & 0 ÷ 0.94 &  0.41 ÷ 1 &  1 ÷ 3.2 &  0 ÷ 0.18 &  -55 ÷ 180 &  -8.3 ÷ 45 &  110 ÷ 380 \\
SQAD &  14 ÷ 170 &  4 ÷ 33 &  0.1 ÷ 0.92 &  0.67 ÷ 1 &  1.4 ÷ 3.8 &  0 ÷ 0.073 &  -140 ÷ 150 &  -4 ÷ 69 &  150 ÷ 450 \\
\hline
\multicolumn{10}{c}{Mean values for Bloom's Taxonomy levels in RTD dataset}\\
Knowledge        & 47 ± 19                                                    & 9.1 ± 3.4                                             & 0.5 ± 0.14                                                   & 0.98 ± 0.047          & 2 ± 0.3             & 0.0041 ± 0.0099    & 34 ± 36          & 20 ± 9           & 250 ± 37     \\
Comprehension    & 65 ± 34                                                    & 12 ± 6.5                                              & 0.44 ± 0.13                                                  & 0.96 ± 0.067          & 2.3 ± 0.42          & 0.0059 ± 0.01    & 8.5 ± 42         & 28 ± 13          & 270 ± 36        \\
Application      & 56 ± 24                                                    & 10 ± 4.2                                              & 0.47 ± 0.13                                                  & 0.97 ± 0.052          & 2.2 ± 0.34          & 0.0049 ± 0.011       & 22 ± 36          & 24 ± 10          & 260 ± 37        \\
Analysis         & 64 ± 32                                                    & 11 ± 5.6                                              & 0.48 ± 0.14                                                  & 0.97 ± 0.058          & 2.3 ± 0.42          & 0.0039 ± 0.0089     & 4.9 ± 44         & 28 ± 14          & 270 ± 40      \\
Synthesis        & 59 ± 38                                                    & 11 ± 6.8                                              & 0.49 ± 0.15                                                  & 0.97 ± 0.06           & 2.2 ± 0.48          & 0.0056 ± 0.012     & 20 ± 49          & 25 ± 15          & 250 ± 38            \\
Evaluation       & 62 ± 31                                                    & 12 ± 5.5                                              & 0.51 ± 0.13                                                  & 0.97 ± 0.058          & 2.3 ± 0.41          & 0.0055 ± 0.013     & 18 ± 44          & 26 ± 13          & 260 ± 43       \\ 
\hline
\multicolumn{10}{c}{Mean values for Bloom's Taxonomy levels in QAJD dataset}\\
Knowledge        & 44 ± 14          & 9.4 ± 2.7         & 0.59 ± 0.15         & 0.97 ± 0.053          & 2.1 ± 0.29          & 0.0056 ± 0.012       & 57 ± 28          & 16 ± 6.8        & 220 ± 30        \\
Comprehension    & 53 ± 14          & 11 ± 2.9          & 0.58 ± 0.13         & 0.98 ± 0.051          & 2.3 ± 0.28          & 0.0038 ± 0.0095         & 45 ± 26          & 20 ± 6.6        & 230 ± 28             \\
Application      & 52 ± 15          & 11 ± 2.9          & 0.58 ± 0.13         & 0.98 ± 0.047          & 2.3 ± 0.28          & 0.0042 ± 0.0094       & 48 ± 28          & 19 ± 7.1        & 230 ± 29        \\
Analysis         & 53 ± 14          & 11 ± 2.7          & 0.61 ± 0.13         & 0.96 ± 0.058          & 2.3 ± 0.25          & 0.0067 ± 0.012          & 44 ± 27          & 20 ± 6.9        & 230 ± 32  \\
Synthesis        & 51 ± 15          & 11 ± 3.2          & 0.61 ± 0.14         & 0.97 ± 0.057          & 2.3 ± 0.28          & 0.0057 ± 0.011          & 54 ± 25          & 18 ± 6.6        & 220 ± 29   \\
Evaluation       & 47 ± 15          & 10 ± 3            & 0.59 ± 0.14         & 0.97 ± 0.052          & 2.2 ± 0.29          & 0.0052 ± 0.013          & 56 ± 27          & 17 ± 6.9        & 220 ± 30       \\ \hline
\end{tabular}
\addtolength{\tabcolsep}{-3pt}  
\caption{Mean values and ranges of complexity measures for explored datasets: (RTD) R. Tatman's Question-Answer Dataset, (QAJD) Question-Answer Jokes, (SQAD) Stanford Question Answering Dataset}
\label{tab:e1_measures_vs_blooms}
\end{sidewaystable}

\subsection{Experiment 2: Answer-Based Assessment}
In this experiment, we measured questions, aggregating their answers, as well as their context.

\subsubsection{Distribution of metrics values.}
\begin{figure}[htp]
  \centering
  \includegraphics[width=\textwidth]{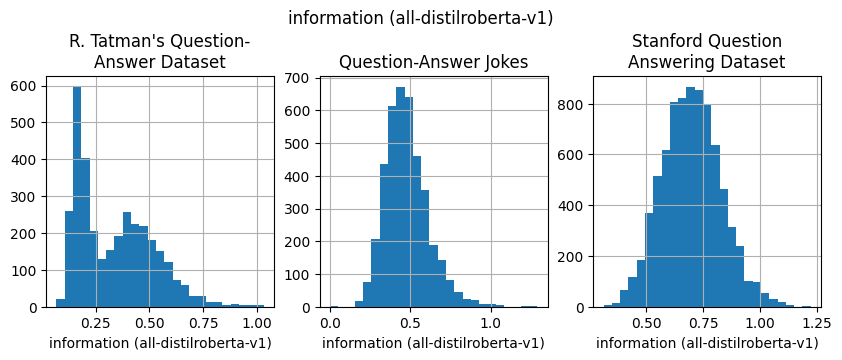}
  \caption{Distribution of the information metric (all-distilroberta-v1 sentence embedder)}
  \label{e2_inf1}
\end{figure}

\begin{figure}[htp]
    \centering
    \includegraphics[width=\textwidth]{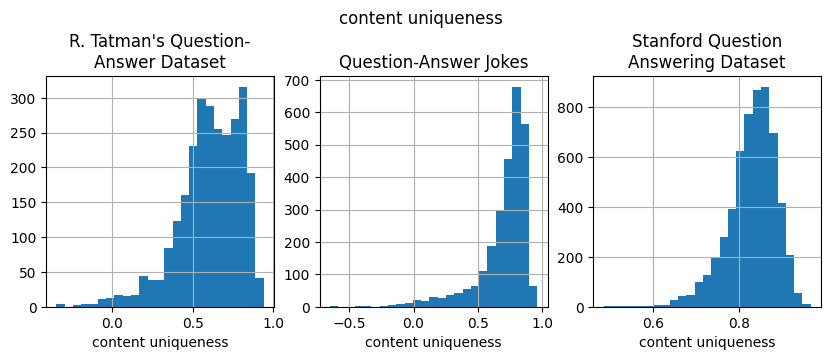}
    \caption{Distribution of the content uniqueness metric values}
    \label{e2_cu}
\end{figure}

\begin{figure}[htp]
    \centering
    \includegraphics[width=\textwidth]{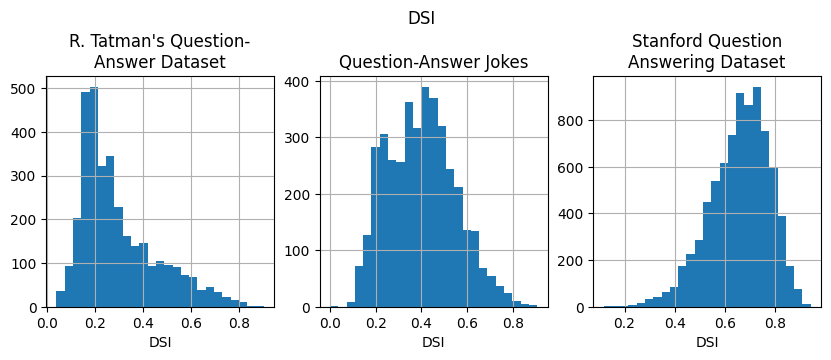}
    \caption{Distribution of the DSI metric values}
    \label{e2_dsi}
\end{figure}

\begin{figure}[htp]
    \centering
    \includegraphics[width=\textwidth]{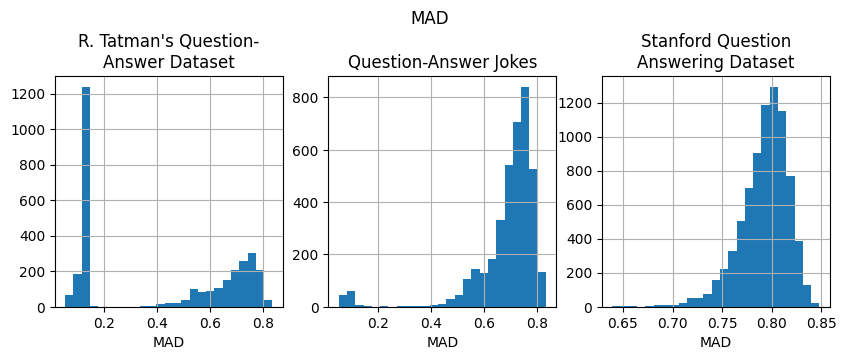}
    \caption{Distribution of the MAD metric values}
    \label{e2_mad}
\end{figure}

The distributions of the answer-based metrics are shown in Figures~\ref{e2_inf1} to \ref{e2_mad}. The overall distribution is similar across datasets, with only slight variations in Question-Answer Jokes dataset, which shows a higher concentration of information metric values around 0.5. R. Tatman’s dataset shows some outliers in the DSI metric, where many questions have DSI values around 0.1.

\subsubsection{Analysis related to the Bloom’s Taxonomy levels.}
Table~\ref{e2_meanstd_tatman_answer} shows the mean and the standard deviation of the answer-based metrics -- information with utilized three embedding models, content uniqueness (CU), DSI, and MAD -- for each Bloom’s Taxonomy level. Here, knowledge questions scored the highest values for the information and the content uniqueness metrics. DSI and MAD did not show significant differences between levels.
\begin{table}[htp]
\resizebox{\textwidth}{!}{
\begin{tabular}{lcccccc}
\hline
& inf\_droberta & inf\_minilm & inf\_mpnet & CU & DSI & MAD \\
\textbf{overall} & \textbf{0.34 ± 0.17} & \textbf{0.41 ± 0.18} & \textbf{0.41 ± 0.17} & \textbf{0.29 ± 0.16} & \textbf{0.41 ± 0.29} & \textbf{0.61 ± 0.2} \\ \hline
Knowledge & 0.37 ± 0.18 & 0.44 ± 0.18 & 0.44 ± 0.17 & 0.32 ± 0.16 & 0.45 ± 0.29 & 0.6 ± 0.21 \\
Comprehension & 0.25 ± 0.13 & 0.32 ± 0.13 & 0.33 ± 0.14 & 0.24 ± 0.14 & 0.34 ± 0.29 & 0.63 ± 0.16 \\
Application & 0.29 ± 0.15 & 0.37 ± 0.16 & 0.38 ± 0.15 & 0.26 ± 0.14 & 0.37 ± 0.29 & 0.59 ± 0.17 \\
Analysis & 0.3 ± 0.17 & 0.37 ± 0.18 & 0.38 ± 0.17 & 0.28 ± 0.17 & 0.4 ± 0.3 & 0.62 ± 0.16 \\
Synthesis & 0.26 ± 0.14 & 0.33 ± 0.14 & 0.35 ± 0.14 & 0.24 ± 0.14 & 0.31 ± 0.27 & 0.62 ± 0.16 \\
Evaluation & 0.29 ± 0.17 & 0.34 ± 0.17 & 0.35 ± 0.16 & 0.25 ± 0.16 & 0.35 ± 0.29 & 0.62 ± 0.18 \\ \hline
\end{tabular}
}
\caption{Results of analysis in terms of Bloom's Taxonomy on R. Tatman's Question-Answer Dataset -- average values of answer-based metrics}
\label{e2_meanstd_tatman_answer}
\end{table}

\begin{figure}[htp]
    \centering
    \includegraphics[width=0.75\textwidth]{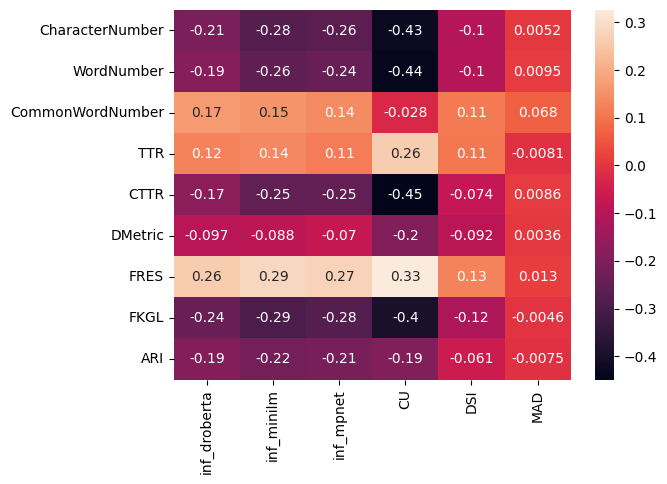}
    \caption{Question-based metrics (from Experiment 1) correlation with Question-with-answer context metrics (from Experiment 2) on R. Tatman's Question-Answer Dataset}
    \label{e2_tatman_corr_table}
\end{figure}

Figure~\ref{e2_tatman_corr_table} presents correlations between the answer-based metrics and the question-only measurements, from the Taxonomy Assessment experiment for R. Tatman's Question-Answer Dataset. 

\begin{figure}[htp]
    \centering
    \includegraphics[width=0.75\textwidth]{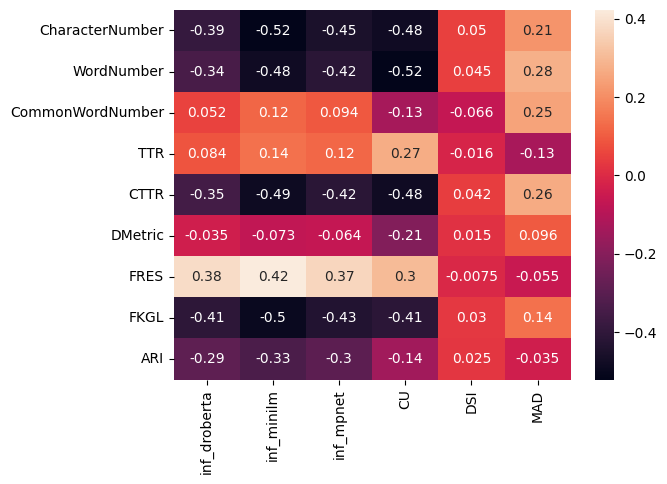}
    \caption{Question-based metrics (from Experiment 1) correlation with Question-with-answer context metrics (from Experiment 2) on Stanford Question Answering Dataset}
    \label{e2_squad_corr_table}
\end{figure}

DSI and MAD are not correlated to any question complexity measure. The information (in all three implementation versions) and the content uniqueness metrics are correlated to some of them, but loosely. Loose correlations might have to do with the definitions of the measures, which are based on cosine distance or on bigrams. Here, a shorter question is more likely to have its context enriched when combined with the answer.
\begin{table}[htp]
\resizebox{\textwidth}{!}{
\begin{tabular}{p{2cm}p{3.5cm}p{4.5cm}cccccc}
\hline
Dataset / Bloom's level & Question & Answer    & inf\_droberta& inf\_minilm & inf\_mpnet& CU & DSI & MAD\\
\hline
\begin{tabular}[c]{@{}l@{}}RTD   \\ Knowledge  \end{tabular}   & What is Antwerp?    & Antwerp is a city and municipality in Belgium      & 0,533    & 0,746                                    & 0,546                                   & 0,714                           & 0,681                            & 0,727                            \\
RTD / Knowledge     & What is Jakarta?       & Indonesia's special capital region              & 0,572                                      & 0,580                                    & 0,479                                   & 0,600                           & 0,695                            & 0,686                            \\
RTD / Analysis      & How does the distribution size of the leopard compare to the distribution of other wild cats? & As of 1996, the leopard had the largest distribution of any wild cat, although populations before and since have shown a declining trend and are fragmented outside of Subsaharan Africa. & 0,587                                      & 0,787                                    & 0,680                                   & 0,627                           & 0,805                            & 0,777                            \\
QAJD / Comprehension & If an ant smells bad, what is it called?                                                      & De-oder-ant                                 & 0,488                                      & 0,470                                    & 0,468                                   & 0,154                           & 0,668                            & —                                \\
QAJD / Analysis      & What's the difference between an Irish wedding and an Irish funeral?                          & One guest                                & 0,251                                      & 0,290                                    & 0,308                                   & 0,188                           & 0,594                            & 0,864                            \\
QAJD  / Evaluation    & Why is it hard to watch two elephants boxing?                                                 & Because they've got the same color trunks      & 0,499                                      & 0,477                                    & 0,621                                   & 0,500                           & 0,779                            & 0,837                            \\
QAJD  / Evaluation    & Why is the long-term liability sad?                                                           & Because it is aloan                       & 0,537                                      & 0,377                                    & 0,244                                   & 0,385                           & 0,714                            & —                                \\
SQUAD / Application   & Why should trees not be planted on the side of a building facing the equator?                 & In climates with significant heating loads, deciduous trees should not be planted on the equator facing side of a building because they will interfere with winter solar availability.    & 0,523                                      & 0,437                                    & 0,554                                   & 0,489                           & 0,796                            & 0,797                            \\
SQUAD / Evaluation    & Why is a second meter usually unnecessary to monitor electricity use?                         & Most standard meters accurately measure in both directions, making a second meter unnecessary.                                                                                            & 0,380                                      & 0,414                                    & 0,539                                   & 0,500                           & 0,780                            & 0,877                          \\ 
SQUAD / Synthesis     & What can happen if people are exposed to antibiotics at a young age?                          & Exposure to antibiotics early in life is associated with increased body mass in humans and mouse models.                                                                                  & 0,658                                      & 0,660                                    & 0,715                                   & 0,545                           & 0,777                            & 0,836                            \\
\hline
\end{tabular}
}
\caption{Examples of questions and answers and their answer-based metrics: (RTD) R. Tatman's Question-Answer Dataset, (QAJD) Question-Answer Jokes, (SQAD) Stanford Question Answering Dataset}
\label{tab:e2_examples}
\end{table}

Analyzing the examples given in Table~\ref{tab:e2_examples} one can see that the relatively high metric values occurred among the questions in which the answer is much longer than the question, e.g. for the short question: ``What is ...?''.

\subsection{Experiment 3: Topic Study}

In this experiment, we analyzed if there are any significant differences, within our metrics, between topics covered in the datasets. Here, we tried topic discovery methods: LDA or SeedTopicMine. In what follows, we provide results for R. Tatman’s Question-Answer Dataset, 
because there are no significant differences between different topics and their configurations with regard to our metrics. 

\begin{figure}[htp]
  \centering
  \includegraphics[width=0.6\textwidth]{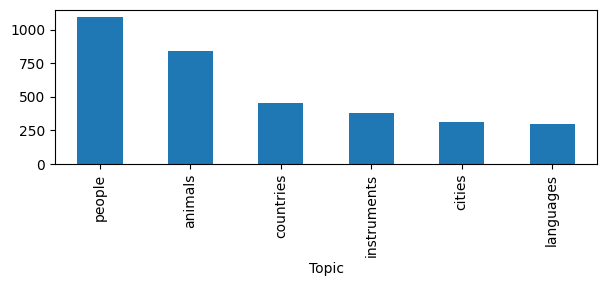}
  \caption{Topic distribution in R. Tatman’s Dataset}
  \label{e3_tatman_topic_hist}
\end{figure}

Figure~\ref{e3_tatman_topic_hist} shows the overall distribution of topics in R. Tatman’s Question-Answer Dataset. Questions in the Knowledge category are dominant across all topics, ranging from 58.8\% in one topic -- ``people'' to 70.6\% in the ``languages'' topic. 
Analysis and Application questions are more common in non-knowledge categories, with the Analysis level prevalent in the ``countries'' and ``languages'' topics.

Table~\ref{e3_meanstd_tatman_answer} displays the mean and the standard deviation of answer-based metrics, across different topics. No significant differences in the information, content uniqueness, DSI and MAD metrics were observed across topics, though questions about ``cities'' scored the highest information values on average. Possibly, this group of questions had more extended answers in the dataset. 
\begin{table}[htp]
\resizebox{\textwidth}{!}{
\begin{tabular}{lcccccc}
\hline
             & inf\_droberta   & inf\_minilm   & inf\_mpnet   & CU          & DSI         & MAD         \\
\textbf{overall} & \textbf{0.34 ± 0.17} & \textbf{0.41 ± 0.18} & \textbf{0.41 ± 0.17} & \textbf{0.29 ± 0.16} & \textbf{0.41 ± 0.29} & \textbf{0.61 ± 0.2} \\ \hline
animals     & 0.33 ± 0.16    & 0.41 ± 0.18  & 0.42 ± 0.17 & 0.32 ± 0.17 & 0.43 ± 0.29 & 0.59 ± 0.19 \\
cities      & 0.37 ± 0.18    & 0.47 ± 0.2   & 0.45 ± 0.17 & 0.31 ± 0.16 & 0.47 ± 0.29 & 0.66 ± 0.19 \\
countries   & 0.32 ± 0.16    & 0.4 ± 0.17   & 0.41 ± 0.16 & 0.29 ± 0.16 & 0.37 ± 0.28 & 0.63 ± 0.18 \\
instruments & 0.33 ± 0.17    & 0.38 ± 0.18  & 0.39 ± 0.17 & 0.28 ± 0.14 & 0.43 ± 0.29 & 0.59 ± 0.21 \\
languages   & 0.32 ± 0.16    & 0.4 ± 0.18   & 0.4 ± 0.17  & 0.29 ± 0.16 & 0.44 ± 0.3  & 0.69 ± 0.15 \\
people      & 0.35 ± 0.19    & 0.41 ± 0.18  & 0.41 ± 0.17 & 0.28 ± 0.16 & 0.39 ± 0.29 & 0.57 ± 0.21 \\ \hline
\end{tabular}
}
\caption{Results of analysis in terms of question topic on R. Tatman's Question-Answer Dataset -- average values of answer-based metrics}
\label{e3_meanstd_tatman_answer}
\end{table}

\subsection{Experiment 4: Interview Study} 
\begin{figure}[htp]
    \centering
    \includegraphics[width=0.75\textwidth]{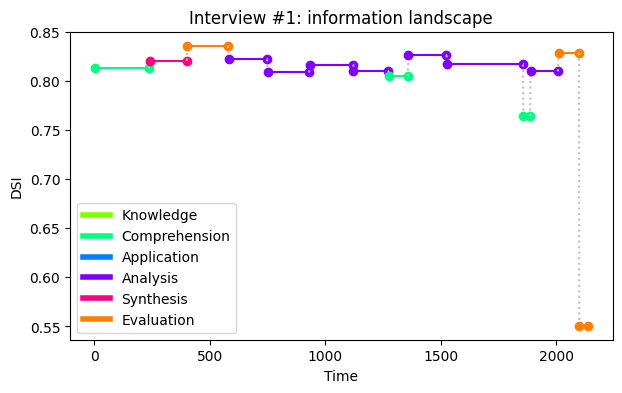}
    \caption{Results of timestamp study in STA interview \#1}
    \label{e4_1}
\end{figure}

\begin{figure}[htp]
    \centering
    \includegraphics[width=0.75\textwidth]{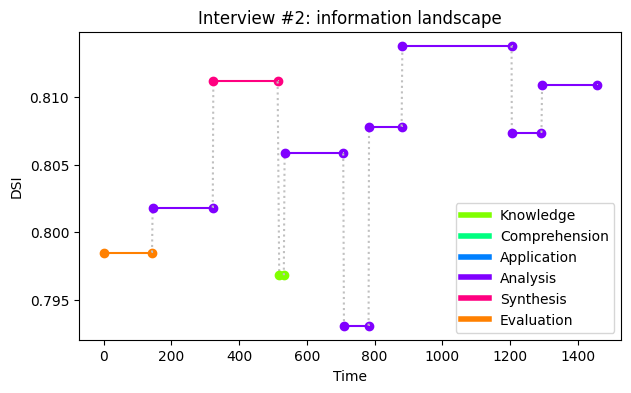}
    \caption{Results of timestamp study in STA interview \#2}
    \label{e4_2}
\end{figure}

\begin{figure}[htp]
    \centering
    \includegraphics[width=0.75\textwidth]{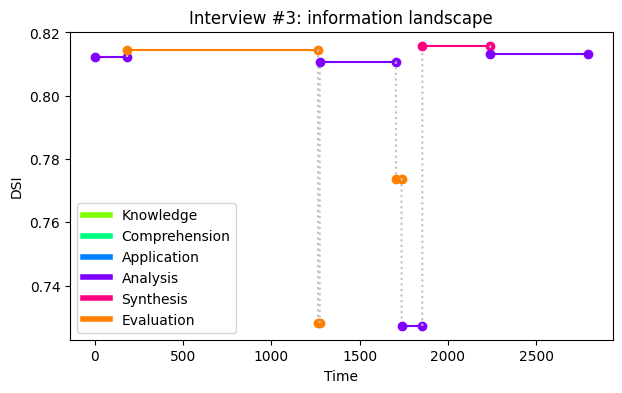}
    \caption{Results of timestamp study in STA interview \#3}
    \label{e4_3}
\end{figure}

\begin{figure}[htp]
    \centering
    \includegraphics[width=0.75\textwidth]{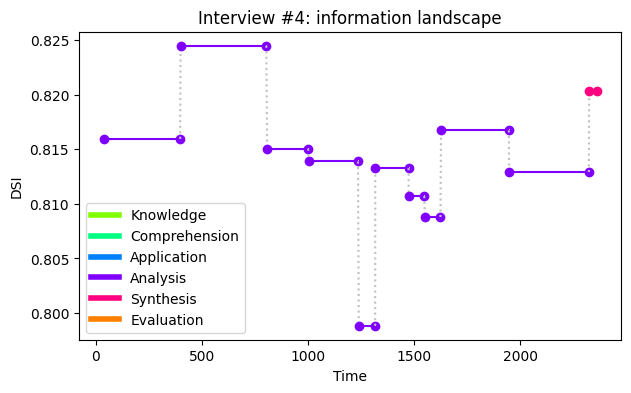}
    \caption{Results of timestamp study in STA interview \#4}
    \label{e4_4}
\end{figure}

\begin{figure}[htp]
    \centering
    \includegraphics[width=0.75\textwidth]{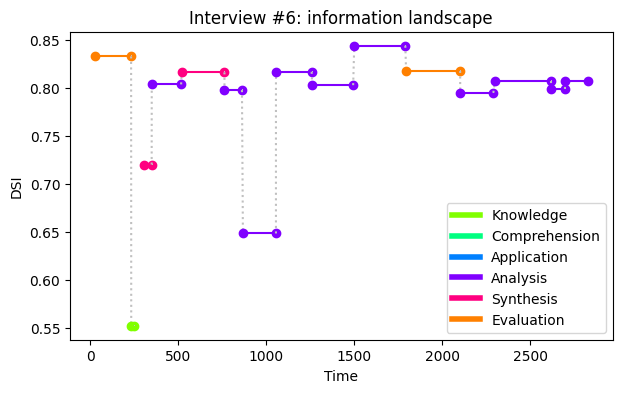}
    \caption{Results of timestamp study in STA interview \#6}
    \label{e4_6}
\end{figure}

In this experiment, we analyzed how the measures and Bloom's taxonomy level change during the flow of interviews -- the sequence of questions and answers. Figures~\ref{e4_1} to \ref{e4_6} display the results of the interview study across the most characteristic STA interviews. 
In most interviews, there was no significant pattern indicating that information depends on the time a question was asked. However, in interview \#2, the DSI values increased toward the end of the interview. Nevertheless, Bloom's Taxonomy levels provide interesting feedback about the interviews (they are marked with colors in the figures); showing that most interviews analyze, synthesize, and evaluate a common topic of the particular interviews.

Overall, typically, short answers reduced the creativity scores, while longer answers that significantly changed the conversation context scored higher. However, DSI (Divergent Semantic Integration) and MAD (Mean Answer Divergence) metrics were less sensitive to variations in question complexity.

\section{Discussion of experimental results}\label{sec:discussion}

Let us now summarize the results and insights gained from the experiments and discuss how they address the posed research questions.

\subsection{RQ1: Can the creativity of questions be measured with NLP techniques?}
In this work, two concepts relating questions to creativity have been considered, one based on using existing ideas to generate new ones, and the second based on~\cite{sonophilia,inquiry}, where a creative question produces a contextually diverse answer, such as deeper analysis. 

The Taxonomy Assessment experiment applied Bloom's Taxonomy, as a tool for creativity measurement. 
The Answer-Based Assessment focused on metrics interpreting creativity as the contextual difference between questions and answers, i.e. Information, Content Uniqueness (CU), Divergent Semantic Integration (DSI), and Maximal Associative Distance (MAD).
Results, including simple examples (see Figure~\ref{e1_examples_lime1}), show that (at least some aspects of) creativity can be assessed using the proposed NLP-based approach. Patterns of question structure, like yes/no questions being classified as Knowledge and comparison-based questions classified as Analysis, suggest that question wording can influence creativity classification. Still, the best and most straightforward indicator of creativity is Bloom's Taxonomy classification. This task gained very high results in the Taxonomy Assessment Experiment. For the other metrics, i.e. lexical diversity, readability and even semantic DSI and MAD, the concept of creativity is too complex to capture its idea and understanding.

\subsection{RQ2: How can one measure if, and how, various types, or characteristics, of a question influence creativity?}

Although the topic and sequencing of questions and answers (timing within the interview) had no measurable effect on creativity, the differing results between Taxonomy and Answer-Based Assessments indicate that combining multiple metrics could be valuable in understanding what influences creativity. The diverse nature of questions suggests that their creativity may still be a complex property to quantify, even for humans. 

To answer RQ3, one can measure creativity by examining broader contextual factors. While topic and timing did not influence creativity, other factors may warrant further exploration in future studies. 
Still, as for now, observing the interview flow, or other metrics, one can conclude that Bloom's Taxonomy is the most promising assessment of question creativity that leads to a more sophisticated and higher level of understanding of the world.

\subsection{RQ3: How does the question lead to a more creative answer?}

%
Analysis, on the basis of Taxonomy Assessment (see Table~\ref{tab:e1_measures_vs_blooms} and Figure~\ref{e1_examples_lime1}), indicates that simpler questions tend to fall under the knowledge category, but longer, complex questions do not necessarily lead to more creative answers. For instance, the longest example question was classified at the Comprehension level, while questions with high character counts often appeared at the Comprehension and Analysis levels.


The answer-based creativity metrics used different methods to assess creativity: Information compared question-answer concatenation, CU measured the percentage of unique word pairs, DSI calculated pair distances, and MAD found the most semantically distant word. Obtained results revealed little correlation between Bloom’s Taxonomy and answer-based metrics. For example, the Knowledge-level question “What is Santiago?” scored high on all answer-based metrics, showcasing how answer content can significantly differ in creativity, regardless of the taxonomy level. Simply said, for short questions, longer answers increase the value of the metrics (see Table~\ref{tab:e2_examples}). Still, this is not directly correlated with creativity.


The next two exploratory experiments studied the impact of topic and occurrence in a dialogue sequence. 
Metrics and Bloom’s Taxonomy distributions across topics were consistent in the Topic Study, and no clear patterns were identified regarding when the specific questions were asked during interviews in the Interview Study.
In summary, a question leads to more creative answers when it allows for broader, more encompassing responses, independent of its topic or timing.

In summary, some aspects of the creativity of questions can be measured with NLP techniques. At this time, the most suitable metric is Bloom's Taxonomy level classification. However, the answer-based metrics can also be potentially useful in this task. Here, nevertheless, it is hard to choose the most reliable one, and they would be able to reliably assess only the most evident examples (see Table~\ref{tab:e2_examples}). The difference between question complexity and creativity is not strict, and this aspect still needs to be explored.

\section{Concluding remarks}\label{sec:conclusions}

This research aimed to determine whether question creativity can be measured using NLP techniques, and how questions influence creative answers. Bloom’s Taxonomy and answer-based metrics were tested in four experiments. The first two experiments evaluated the creativity measures (Bloom's Taxonomy and answer-based metrics) by comparing them with question statistics and by analyzing simple examples. The last two experiments were exploratory. The Topic Study related these measures with topic modelling, while the Interview Study examined the metrics in interviews.

The tools performed well, particularly on the provided examples. Results suggested that metric values primarily depend on question length and content, although no direct correlation was found with the text complexity, topic, or occurrence in a dialogue sequence. This can be explained by the diversity of questions.

Several challenges and limitations were present in this research. Thematically, the task combined data science and psychology, making it complex. Since question creativity lacks a standard definition, preliminary assumptions were required. Additionally, there were no datasets specifically tailored to conduct creativity-related research. Moreover, most of the data used consisted of factual questions, which typically do not lead to creative answers. Technically, while Bloom’s Taxonomy classification model was accurate, there is still room for improvement. Future work could benefit from datasets specifically focused on the cognitive potential and on the creativity in questions. Another very promising direction, of further research, is utilizing the potential of large language models, e.g. GPT-4, LLama, with elaborated prompts that lead the models to ``reason on questions and answers'' to assess their complexity, diversity, coherence, and also creativity. 

Beyond question measuring itself, the question analysis might also support automatic question generation tasks, which can reduce manual question creation costs and meet the growing demand for questions, particularly in education and in dialogue systems \cite{aqg,chatbots}.

\section*{Acknowledgments}
The authors of this work were funded by the European Union under the Horizon Europe grant OMINO – Overcoming Multilevel INformation Overload (grant number 101086321, http://ominoproject.eu). Views and opinions expressed are those of the authors alone and do not necessarily reflect those of the European Union or the European Research Executive Agency. Neither the European Union nor the European Research Executive Agency can be held responsible for them. A.W. was also co-financed with funds from the Polish Ministry of Education and Science under the programme entitled International Co-Financed Projects.

%
%
%

\begin{thebibliography}{100} 

\bibitem{cambridge} \emph{question}, 2023, in \url{dictionary.cambridge.org}. Retrieved December 8, 2023, from \url{https://dictionary.cambridge.org/us/dictionary/english/question}.
\bibitem{differentTypes} \emph{Different Types of Questions}, Revolution Learning and Development. Retrieved December 9, 2023, from \url{https://www.revolutionlearning.co.uk/article/different-types-of-questions/}.
\bibitem{15ways} J.G. Miller, \emph{15 Reasons to Ask Questions}, QBQ, 2014. Retrieved December 8, 2023, from \url{https://qbq.com/15-reasons-to-ask-questions/}.
\bibitem{eperel} E. Perel, M.A. Miller, \emph{Letters from Esther \#52: A Good Question Changes the Story}, Esther Perel's Blog. Retrieved February 26, 2024, from \url{https://www.estherperel.com/blog/letters-from-esther-52-a-good-question-changes-the-story}.
\bibitem{akinator} G. Sasson, Y.N. Kenett, \emph{A Mirror to Human Question Asking: Analyzing the Akinator Online Question Game}, Big Data and Cognitive Computing, 2023; 7(1):26. \textsc{doi}: \texttt{10.3390/bdcc7010026}
\bibitem{inquiry} G. Sasson, T. Raz, Y.N. Kenett, \emph{The Art of Creative Inquiry—From Question Asking to Prompt Engineering}, Journal of Creative Behaviour, 2024. \textsc{doi}: \texttt{10.1002/jocb.671}.
\bibitem{aqt} T. Raz, R. Reiter-Palmon, Y.N. Kenett, \emph{The role of asking more complex questions in creative thinking}, Psychology of Aesthetics, Creativity, and the Arts, 2023. \textsc{doi}: \texttt{10.1037/aca0000658}.
\bibitem{raz2024bridging}
  T. Raz, S. Luchini, R Beaty, Y. Kenett,
  \emph{Bridging the Measurement Gap: a Large Language Model Method of Assessing Open-Ended Question Complexity}, in: 
  {Proceedings of the Annual Meeting of the Cognitive Science Society}, 46, 2024
\bibitem{acar2023}
S. Acar, \emph{Creativity Assessment, Research, and Practice in the Age of Artificial Intelligence}, in: {Creativity Research Journal}, 2023, \url{https://doi.org/10.1080/10400419.2023.2271749}
\bibitem{mum1997}
M. Mumford, D. L. Whetzel, R. Reiter-Palmon,
\emph{Thinking Creatively at Work: Organization Influences on Creative Problem Solving}, {The Journal of Creative Behavior},
31(1): 7--17, 1997, \url{https://onlinelibrary.wiley.com/doi/abs/10.1002/j.2162-6057.1997.tb00777.x}
\bibitem{reiter1998}
R. Reiter-Palmon, M. D. Mumford, K. V. Threlfall, \emph{Solving Everyday Problems Creatively: The Role of Problem Construction and Personality Type}, {Creativity Research Journal}, 11(3):187--197, 1998
\bibitem{tofade2013}
T. Tofade, J. Elsner, S. T. Haines,
\emph{Best Practice Strategies for Effective Use of Questions as a Teaching Tool}, {American Journal of Pharmaceutical Education}, 77(7), 2013, 
\url{https://www.sciencedirect.com/science/article/pii/S0002945923029911}
\bibitem{adams2015conducting}
  W.C. Adams, \emph{Conducting semi-structured interviews}, {Handbook of practical program evaluation},
  pp.{492--505}, 2015, {Wiley Online Library}
\bibitem{daud2012}
{A.M. Daud, J. Omar, P. Turiman, K.  Osman}, \emph{Creativity in Science Education}, {Procedia - Social and Behavioral Sciences},
59: 467-474, 2012
\url{https://doi.org/10.1016/j.sbspro.2012.09.302}
\bibitem{kenett2024assessing}
  Y.N. Kenett, \emph{Assessing the role of associative abilities in creative thinking via behavioral, computational, and neuroscientific approaches}, {Handbook of Creativity Assessment}, pp.{182--198}, {2024}, {Edward Elgar Publishing}
\bibitem{taxonomy} L.W. Anderson, D.R. Krathwohl, B.S. Bloom, \emph{A Taxonomy for Learning, Teaching, and Assessing: A Revision of Bloom's Taxonomy of Educational Objectives}, Longman, New York, 2000. \textsc{isbn}: 978-0-8013-1903-7.
\bibitem{stems} J. Rutka, \emph{Bloom’s Taxonomy Question Stems for Use in Assessment}, Top Hat, 2023. Retrieved December 9, 2023, from \url{https://tophat.com/blog/blooms-taxonomy-question-stems}.
\bibitem{taxonomyWebpage} \emph{Bloom’s Taxonomy}. Retrieved December 9, 2023, from \url{https://www.bloomstaxonomy.net}.
\bibitem{mrst} Sara Trickey, \emph{Bloom’s Taxonomy}, Teach With Mrs T, 2020. Retrieved December 18, 2023, from \url{https://www.teachwithmrst.com/post/bloom-s-taxonomy}.
\bibitem{krathwohl} D.R. Krathwohl, B.S. Bloom, B.B. Masia, \emph{Taxonomy of educational objectives: Handbook II: Affective domain}, David McKay Co., New York, 1964.
\bibitem{simpson} E.J. Simpson, \emph{The Classification of Educational Objectives, Psychomotor Domain}, 1972.
\bibitem{sonophilia} \emph{Dr. Yoed Kenett – The brain is wired to adapt, survive, and evolve in complex environments}. An interview with Y.N. Kenett, conducted by Sonophilia Foundation, 2022. Retrieved December 10, 2023, from {\small https://www.sonophiliafoundation.org/dr-yoed-kenett-the-brain-is-wired-to-adapt-survive-and-evolve-in-complex-environments/}
\bibitem{chatbots} H. Shum, X. He, D. Li, \emph{From eliza to xiaoice: 
challenges and opportunities with social chatbots}, Frontiers of Information Technology \& Electronic Engineering, 19(1), 2018, 10-–26.
\bibitem{wtc} Z. Xu, A. Howarth, N. Briggs, N. Cristianini, \emph{What makes us curious? Analysis of a corpus of open-domain questions}, 2021. arXiv: \texttt{2110.15409}. \textsc{url}: \url{https://arxiv.org/abs/2110.15409}
\bibitem{whatif} \emph{The story of Project What If}, We the Curious. Retrieved Decemebr 14, 2023, from \url{https://www.wethecurious.org/projectwhatif}.
\bibitem{ecuador} D. Buenaño-Fernandez, M. González, D. Gil, S. Luján-Mora, \emph{Text Mining of Open-Ended Questions in Self-Assessment of University Teachers: An LDA Topic Modeling Approach}, IEEE Access, vol. 8, 2020, 35318--35330 \textsc{doi}: \texttt{10.1109/ACCESS.2020.2974983}.
\bibitem{examq} M. O. Gani, R. K. Ayyasamy, A. Sangodiah, Y. T. Fui, \emph{Bloom’s Taxonomy-based exam question classification: The outcome of CNN and optimal pre-trained word embedding technique}, 2023. \textsc{doi}: \texttt{10.1007/s10639-023-11842-1}
\bibitem{examqds} M. O. Gani, A. Sangodiah, \emph{Exam Question Datasets}, Version 3, 2023. \textsc{url}: \url{https://figshare.com/articles/dataset/Exam_Question_Datasets/22597957/3}
\bibitem{eq1} A. A. Yahya, A. Osman, A. Taleb, A. A. Alattab, \emph{Analysing the cognitive level of classroom questions using machine learning techniques}, Procedia - Social and Behavioral Sciences, 97, 2013, 587--595. \textsc{doi}: \texttt{10.1016/j.sbspro.2013.10.277}
\bibitem{eq2} M. Mohammed, N. Omar, \emph{Question classification based on Bloom's Taxonomy using enhanced TF-IDF International}, Journal on Advanced Science, Engineering and Information Technology, 8(4--2), 2018, 1679--1685. \textsc{doi}: \texttt{10.18517/ijaseit.8.4-2.6835}
\bibitem{eq3} M. Mohammed, N. Omar, \emph{Question classification based on Bloom's taxonomy cognitive domain using modified TF-IDF and word2vec}, PLoS ONE, 15(3), 2020, 1--21. \textsc{doi}: \texttt{10.1371/journal.pone.0230442}
\bibitem{eq4} S. Das, S. K. D. Mandal, A. Basu, \emph{Identification of cognitive learning complexity of assessment questions using multi-class text classification}, Contemporary Educational TEchnology, 12(2), 2020, 1--14. \textsc{doi}: \texttt{10.30935/cedtech/8341}
\bibitem{eq5} S. Shaikh, S. M. Daudpotta, A. S> Imran, \emph{Bloom's learning outcomes' automatic classification using LSTM and pretrained word embeddings}, IEEE Access, 9, 2021, 117887--117909. \textsc{doi}: \texttt{10.1109/ACCESS.2021.3106443}
\bibitem{eq6} H. Sharma, R. Mathur. T. Chintala, S. Dhanalakshmi, R. Senthil, \emph{An effective deep learning pipeline for improved question classification into Bloom's Taxonomy's domains}, Education and Information Technologies, 2022, 1--41. \textsc{doi}: \texttt{10.1007/s10639-022-11356-2}
\bibitem{aqg} G. Kurdi, J. Leo, B. Parsia et al., \emph{A Systematic Review of Automatic Question Generation for Educational Purposes}. Int J Artif Intell Educ 30, 2020, 121–-204. \textsc{doi}: \texttt{10.1007/s40593-019-00186-y}
\bibitem{mcq1} J. D. Hansen, L. Dexter, \emph{Quality Multiple-Choice Test Questions: Item-Writing Guidelines and an Analysis of Auditing Testbanks}. Journal of Education for Business, 73(2), 1997, 94–97. \textsc{doi}: \texttt{10.1080/08832329709601623}
\bibitem{mcq2} M. Tarrant, A. Knierim, S. K. Hayes, J. Ware, \emph{The frequency of item writing flaws in multiple-choice questions used in high stakes nursing assessments}. Nurse Education in Practice, 6(6), 2006, 354–-363. \textsc{doi}: \texttt{10.1016/j.nepr.2006.07.002}
\bibitem{mcq3} M. R. Hingorjo, F. Jaleel, \emph{Analysis of one-best MCQs: the difficulty index, discrimination index and distractor efficiency}. JPMA-Journal of the Pakistan Medical Association, 62(2), 2012, 142.
\bibitem{mcq4} B.R. Rush, D.C. Rankin, B.J. White, \emph{The impact of item-writing flaws and item complexity on examination item difficulty and discrimination value}. BMC Med Educ 16, 2016, 250. \textsc{doi}: \texttt{10.1186/s12909-016-0773-3}
\bibitem{ld} P.M. McCarthy, S. Jarvis, \emph{MTLD, vocd-D, and HD-D: A validation study of sophisticated approaches to lexical diversity assessment}, 2010. \textsc{doi}: \texttt{10.3758/BRM.42.2.381}
\bibitem{lexrich} F.J. Tweedie, R.H. Baayen, \emph{How Variable May a Constant be? Measures of Lexical Richness in Perspective}, Computers and the Humanities, 32, Kluwer Academic Publishers, 1998, 323--352.
\bibitem{voceval} O. Vinogradova, \emph{Automated Vocabulary Evaluation in a Learner Corpus}. Polylinguality and Transcultural Practices, 15, 2018, 372-380. \textsc{doi}: \texttt{10.22363/2618-897X-2018-15-3-372-380}. 
\bibitem{ttr} M.C. Templin, \emph{Certain language skills in children; their development and interrelationships}. University of Minnesota Press, 1957.
\bibitem{marketing} M. Reisenbichler, T. Reutterer, D. A. Schweidel, D. Dan, \emph{Frontiers: Supporting Content Marketing with Natural Language Generation}, Marketing Science, 41(3), 2022, 441--452. \textsc{doi}: \texttt{10.1287/mksc.2022.1354}.
\bibitem{dsi} D.R. Johnson et al., \emph{Divergent semantic integration (DSI): Extracting creativity from narratives with distributional semantic modeling}, Behav Res Methods 55(7), October 2022, 3726--3759. \textsc{doi}: \texttt{10.3758/s13428-022-01986-2}.
\bibitem{mad} Y. Yu et al., \emph{A MAD method to assess idea novelty: Improving validity of automatic scoring using maximum associative distance (MAD)}, Psychology of Aesthetics, Creativity, and the Arts, 2023. \textsc{doi}: \texttt{10.1037/aca0000573}.
\bibitem{readable} M. Xia, E. Kochmar, T. Briscoe, \emph{Text Readability Assessment for Second Language Learners}, in: \emph{In Proceedings of the 11th Workshop on Innovative Use of NLP for Building Educational Applications}, Association for Computational Linguistics, San Diego, 2016, 12--22. \textsc{doi}: \texttt{10.18653/v1/W16-0502}.
\bibitem{flesch} R. Flesch, \emph{How to Write Plain English: a Book for Lawyers and Consumers}, University of Canterbury, 1979.
\bibitem{fleschk} J.P. Kincaid et al., \emph{Derivation of new readability formulas (Automated Readability Index, Fog Count and Flesch Reading Ease Formula) for Navy enlisted personnel}, Research Branch Report 8--75, Millington, TN: Naval Technical Training, U. S. Naval Air Station, Memphis, TN, 1975. \textsc{url}: \url{https://apps.dtic.mil/sti/pdfs/ADA006655.pdf}
\bibitem{flesch_refine} A.A. Hussin, \emph{Refining the Flesch Reading Ease Formula for Intermediate and High-Intermediate ESL Learners}, 2015. 
\bibitem{ari} R.J. Senter, E.A. Smith, \emph{Automated Readability Index}, 1967.
\bibitem{coursera} \emph{What is Natural Language Processing?}, Coursera, 2023. Retrieved December 16, 2023, from \url{https://www.coursera.org/articles/natural-language-processing}
\bibitem{w2v} T. Mikolov, K. Chen, G. Corrado, J. Dean, \emph{Efficient Estimation of Word Representations in Vector Space}, 2013. arXiv: \texttt{1301.3781}. \textsc{url}: \url{https://arxiv.org/abs/1301.3781}
\bibitem{glove} M.E. Peters, M. Neumann, M. Iyyer, M. Gardner, C. Clark, K. Lee, L. Zettlemoyer, \emph{Deep contextualized word representations}, 2018. arXiv: \texttt{1802.05365}. \textsc{url}: \url{https://arxiv.org/abs/1802.05365}
\bibitem{elmo} J. Pennington, R. Socher, C. Manning, \emph{GloVe: Global Vectors for Word Representation}, Association for Computational Linguistics D14--1162, 2014, 1532--1543. \textsc{doi}: \texttt{10.3115/v1/D14-1162}
\bibitem{cohere} L. Serrano, \emph{What Are Word and Sentence Embeddings?}, Cohere Blog, 2023. Retrieved December 17, 2023, from \url{https://txt.cohere.com/sentence-word-embeddings/}
\bibitem{transformer} A. Vaswani, N. Shazeer, N. Parmar, J. Uszkoreit, L. Jones, A.N. Gomez, Ł. Kaiser, I. Polosukhin, \emph{Attention Is All You Need}, 2017. arXiv: \texttt{1706.03762}. \textsc{url}: \url{https://arxiv.org/abs/1706.03762}
\bibitem{bert} J. Devlin, M.-W. Chang, K. Lee, K. Toutanova, \emph{BERT: Pre-training of Deep Bidirectional Transformers for Language Understanding}, 2018. arXiv: \texttt{1810.04805}. \textsc{url}: \url{https://arxiv.org/abs/1810.04805}
\bibitem{bert_architecture} S. Gundapu, R. Mamidi, \emph{Transformer based automatic COVID-19 fake news detection system}, 2021. arXiv: \texttt{2101.00180}. \textsc{url}: \url{https://arxiv.org/abs/2101.00180}
\bibitem{transfer} F. Zhuang, Z. Qi, K. Duan, D. Xi, Y. Zhu, H. Zhu, H. Xiong, Q. He, \emph{A Comprehensive Survey on Transfer Learning}, 2019. arXiv: \texttt{1911.02685}. \textsc{url}: \url{https://arxiv.org/abs/1911.02685}
\bibitem{sbert} N. Reimers, I. Gurevych, \emph{Sentence-BERT: Sentence Embeddings using Siamese BERT-Networks}, 2019. arXiv: \texttt{1908.10084}. \textsc{url}: \url{https://arxiv.org/abs/1908.10084}
\bibitem{sbertnet} \emph{SentenceTransformers Documentation}. \textsc{url}: \url{https://sbert.net/index.html}
\bibitem{roberta} Y. Liu, M. Ott, N. Goyal, J. Du, M. Joshi, D. Chen, O. Levy, M. Lewis, L. Zettlemoyer, V. Stoyanov, \emph{RoBERTa: A Robustly Optimized BERT Pretraining Approach}, 2019. arXiv: \texttt{1907.11692}. \textsc{url}: \url{https://arxiv.org/abs/1907.11692}
\bibitem{distilbert} V. Sanh, L. Debut, J. Chaumond, T. Wolf, \emph{DistilBERT, a distilled version of BERT: smaller, faster, cheaper and lighter}, 2019. arXiv: \texttt{1910.01108}. \textsc{url}: \url{https://arxiv.org/abs/1910.01108}
\bibitem{minilm} W. Wang et al., \emph{MiniLM: Deep Self-Attention Distillation for Task-Agnostic Compression of Pre-Trained Transformers}, 2020. arXiv: \texttt{2002.10957}. \textsc{url}: \url{https://arxiv.org/abs/2002.10957}
\bibitem{mpnet} K. Song et al., \emph{MPNet: Masked and Permuted Pre-training for Language Understanding}, 2020. arXiv: \texttt{2004.09297v2}. \textsc{url}: \url{https://arxiv.org/abs/2004.09297v2}
\bibitem{xlnet} Z. Yang et al. \emph{XLnet: Generalized autoregressive pretraining for language understanding}, 2019.
\bibitem{gpt1} A. Radford, K. Narasimhan, \emph{Improving Language Understanding by Generative Pre-Training}, 2018.
\bibitem{gpt2} A. Radford, J. Wu, R. Child, D. Luan, D. Amodei, I. Sutskever, \emph{Language Models are Unsupervised Multitask Learners}, 2019.
\bibitem{gpt3} T. B. Brown et al., \emph{Language models are few-shot learners}, 2020. arXiv: \texttt{2005.14165}. \textsc{url}: \url{https://arxiv.org/abs/2005.14165}
\bibitem{gpt4} OpenAI, \emph{GPT-4 Technical Report}, 2023. arXiv: \texttt{2303.08774}. \textsc{url}: \url{https://arxiv.org/abs/2303.08774}
\bibitem{forbes_gpt} R. Toews, \emph{The Next Generation of Large Language Models}, Forbes, 2023. Retrieved December 18, 2023, from \url{https://www.forbes.com/sites/robtoews/2023/02/07/the-next-generation-of-large-language-models}
\bibitem{forbes_io} D. Newman, \emph{Exploring The Ins And Outs Of The Generative AI Boom}, Forbes, 2023. Retrieved December 18, 2023, from \url{www.forbes.com/sites/danielnewman/2023/03/14/exploring-the-ins-and-outs-of-the-generative-ai-boom}
\bibitem{blei_lafferty} D.M. Blei, J.D. Lafferty, \emph{Topic Models}, Text Mining, Chapman and Hall/CRC, 2009, 101--124. \textsc{url}: \url{https://www.cs.columbia.edu/~blei/papers/BleiLafferty2009.pdf}
\bibitem{levity} S. Sheridan, \emph{What Is Topic Modelling? A Beginner's Guide}, Levity Blog, 2022. Retrieved December 11, 2023, from \url{https://levity.ai/blog/what-is-topic-modeling}.
\bibitem{bioinf_tm} L. Liu, L. Tang, W. Dong, S. Yao, W. Zhou, \emph{An overview of topic modeling and its current applications in bioinformatics}. SpringerPlus 5, 1608, 2016. \textsc{doi}: \texttt{10.1186/s40064-016-3252-8}
\bibitem{popgen_lda} J. K. Pritchard, M. Stephens, P. Donnelly, \emph{Inference of Population Structure Using Multilocus Genotype Data}, Genetics, Volume 155, Issue 2, 1 June 2000, 945--959. \textsc{doi}: \texttt{10.1093/genetics/155.2.945}
\bibitem{lda_ml} D.M. Blei, A.Y. Ng, M.I. Jordan, \emph{Latent Dirichlet Allocation}, Journal of Machine Learning Research 3, 2003, 993--1022. \textsc{url}: \url{https://www.jmlr.org/papers/volume3/blei03a/blei03a.pdf}
\bibitem{ensemble} J. Lee, J. Kang, S. Jun, H. Lim, D. Jang, S. Park, \emph{Ensemble Modeling for Sustainable Technology Transfer}, Sustainability, 10, 2278, 2018. \textsc{doi}: \texttt{10.3390/su10072278}. 
\bibitem{seedtm} Y. Zhang, Y. Zhang, M. Michalski, Y. Jiang, Y. Meng, J. Han, \emph{Effective Seed-Guided Topic Discovery by Integrating Multiple Types of Contexts}, 2023. arXiv: \texttt{2212.06002}. \textsc{url}: \url{https://arxiv.org/abs/2212.06002}
\bibitem{underg} N. A. Smith, M. Heilman, R. Hwa, \emph{Question generation as a competitive undergraduate course project}, Proceedings of the NSF Workshop on the Question Generation Shared Task and Evaluation Challenge, September 2008.
\bibitem{rtatman} R. Tatman, \emph{Question-Answer Dataset}, 2017. Retrieved December 14, 2023, from \url{https://www.kaggle.com/datasets/rtatman/questionanswer-dataset/data}.
\bibitem{jiriroz} J. Roznovjak, \emph{Question-Answer Jokes}, 2017. Retrieved December 14, 2023, from \url{https://www.kaggle.com/datasets/jiriroz/qa-jokes}.
\bibitem{stanfordu} P. Rajpurkar, J. Zhang, K. Lopyrev, P. Liang, \emph{SQuAD: 100,000+ Questions for Machine Comprehension of Text}, 2016. arXiv: \texttt{1606.05250}. \textsc{url}: \url{https://arxiv.org/abs/1606.05250}
\bibitem{neural} Q. Zhou, N. Yang, F. Wei, C. Tan, H. Bao, M. Zhou, \emph{Neural Question Generation from Text: A Preliminary Study}, 2017. arXiv: \texttt{1704.01792}. \textsc{url}: \url{https://arxiv.org/abs/1704.01792}
\bibitem{neuralcode} Code for the paper \emph{Neural Question Generation from Text: A Preliminary Study}. \textsc{url}: \url{https://res.qyzhou.me/redistribute.zip} (download link)
\bibitem{qqp} L. Jiang, M. Risdal, N. Dandekar et al., \emph{Quora Question Pairs}, Kaggle, 2017. Retrieved on December 15, 2023, from \url{https://kaggle.com/competitions/quora-question-pairs}.
\bibitem{lime} M. T. Ribeiro, S. Singh, C. Guestrin, \emph{"Why Should I Trust You?": Explaining the Predictions of Any Classifier}, 2016. arXiv: \texttt{1602.04938}. \textsc{url}: \url{https://arxiv.org/abs/1602.04938}
\end{thebibliography}
%

\end{document}